\documentclass[12pt]{article}

\usepackage {latexsym,graphicx,mypaper,xurl,amssymb,hyperref,authblk}

\renewcommand{\abb}[3]{
    \begin{figure}
    \centerline{
      \includegraphics*[scale = #2]{figs/#1.pdf}
    }
    \caption{#3}
    \label{fig:#1}
    \end{figure}
}

\title{Graph Rewriting for Graph Neural Networks}

\author{Adam Machowczyk}
\author{Reiko Heckel}
\affil{University of Leicester, UK}
\affil[ ]{\textit \{amm106,rh122\}@le.ac.uk}

\newcommand{\ifnonempty}[2]{\ifthenelse{\equal{#1}{}}{}{#2}}

\newcounter{comCounter}[section]

\newcommand{\by}[1]{\ifnonempty{#1}{\textbf{#1:}}}

\newcommand{\margincomment}[2][]{%
  \stepcounter{comCounter}%
  $^{\thecomCounter}$
  \marginpar{\scriptsize\by{#1}$^{\thecomCounter}$ #2}%
}

\newcommand{\commentsoff}{%
  \renewcommand{\margincomment}[2][]{\relax}%
  \let\textcomment\comment%
  \let\endtextcomment\endcomment%
}

\begin{document}

\pagestyle{plain}
\maketitle

\begin{abstract}
Given graphs as input, Graph Neural Networks (GNNs) support the inference of nodes, edges, attributes, or graph properties. Graph Rewriting investigates the rule-based manipulation of graphs to model complex graph transformations. We propose that, therefore, (i) graph rewriting subsumes GNNs and could serve as formal model to study and compare them, and (ii) the representation of GNNs as graph rewrite systems can help to design and analyse GNNs, their architectures and algorithms. Hence we propose Graph Rewriting Neural Networks (GReNN) as both novel semantic foundation and engineering discipline for GNNs.
We develop a case study reminiscent of a Message Passing Neural Network realised as a Groove graph rewriting model and explore its incremental operation in response to dynamic updates.
\end{abstract}

\section{Introduction}

Neural Networks (NNs) are among the most successful techniques for machine learning. Starting out from application data, e.g. in a relational database, NNs require an encoding of the data into discrete vectors. 
Graph Neural Networks (GNNs) operate on graphs instead of tabular data. According to~\cite{sanchez-lengeling_reif_pearce_wiltschko_2021} a GNN is “an optimizable transformation on all attributes of the graph (nodes, edges, global context) that preserves graph symmetries (permutation invariances)”. GNNs are applied to~\cite{sanchez-lengeling_reif_pearce_wiltschko_2021} images, text, molecules, social networks, programming code~\cite{allamanis_brockschmidt_khademi_2018} and even mathematical equations~\cite{lample_charton_2019}, in areas  such as antibacterial discovery~\cite{antibac}, physics simulation~\cite{sanchez-gonzalez_godwin_pfaff_ying_leskovec_battaglia_2020}, fake news detection~\cite{monti_frasca_eynard_mannion_bronstein_2019}, traffic prediction~\cite{lange_perez_2020} and recommendation systems~\cite{eksombatchai_jindal_liu_liu_sharma_sugnet_ulrich_leskovec_2017}. 
They support supervised, semi-supervised or unsupervised learning (where data labelling is expensive or impossible) and allow links between input vectors, resembling more closely the structure of the application data. 
In a recommender system where engagements with social media posts are nodes, they may predict such nodes based on the history of engagement, with attributes representing the strength of engagement.

Graph rewriting is the rule-based transformation of graphs~\cite{heckel_taentzer_2020}. Since GNNs are essentially graph transformations, we ask if they can be modelled by graph rewriting, and what the benefits of such a representation might be. Drawing from its rich theory and tool support, we would like to explore the use of graph rewriting as a computational model for GNNs, to specify and compare GNN variants, study their expressiveness and complexity, and as a design and prototyping language for GNNs, evaluating specific GNN architectures and analysing interactions between GNNs and the software and real-world systems with which they interact.
In modern data-driven applications, data is diverse, large-scale, evolving, and distributed, demanding incremental and distributed solutions. This has led to a variety of GNNs~\cite{ZHOU202057} and a need for a unifying theory to describe their expressiveness and cost~\cite{michaelbeyond2022}, evolution and incremental updates~\cite{DBLP:journals/corr/abs-2006-14422}.

In this short paper,  we suggest that typed attributed graph rewriting with control can address these demands. We will illustrate this claim by means of a case study of a recommender system for posts in a social network modelled as a \emph{Graph Rewriting Neural Network (GReNN)}, a controlled graph rewriting system based on a cycle of training and inference transformations with the possibility of dynamic data updates. Training involves the tuning of parameters such as weights and thresholds to support the inference of graph features or properties.
The example does not follow a particular GNN architecture, but can be seen as an elaboration of message-passing GNNs over directed heterogeneous graphs with dynamic changes and incremental updates, a combination we have not encountered in any existing formalism. In particular, typed attributed graphs allow for heterogeneity, where nodes of different types have different attributes and are connected by differently-typed edges, while the rule-based approach supports both local inference operations of message-passing GNNs and updates to the graph structure representing changes in the real-world data.

\section{A Recommender System GReNN}
We consider a social network with users, posts, and engagements, e.g., reads, likes or retweets, modelled as a Groove graph rewriting system~\cite{DBLP:journals/sttt/GhamarianMRZZ12}. In the type graph 

\centerline{
\includegraphics*[width = .4\textwidth]{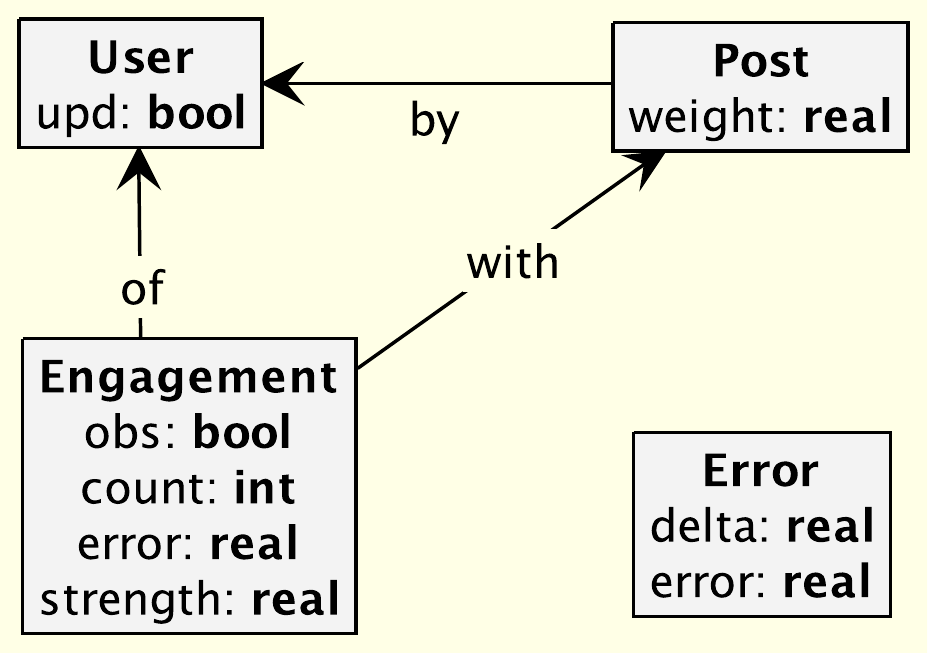}
}

\noindent
users have an \emph{upd} attribute that determines if training is required on engagements with posts by this user.
Posts have \emph{weight}s that model how representative they are of their author, i.e., how well engagements with these posts predict engagements with other posts by that user. Engagements have a \emph{strength} attribute to the model quality of the interaction of a user with a post, e.g. a read may be weaker than a like or retweet. Engagements also have \emph{error, obs} and \emph{count} attributes. The error represents the difference between the actual strength of engagement and the strength inferred from engagements with similar posts. This is used for training the posts’ weights by the gradient descent rule. The Boolean \emph{upd} and \emph{obs} attributes signal, respectively, if an update is required to the training after a change of data affecting this node and if the strength of the node is observed from real-world engagement as opposed to being inferred. The \emph{count} attribute, which is always 1, is used to calculate the cardinality of Engagement nodes in a multi-pattern match in Groove. 
To control training we add an Error node with \emph{error} and \emph{delta} attributes. All attributes except \emph{strength} are part of the general mechanism while \emph{weight} also has application-specific meaning.

\abb{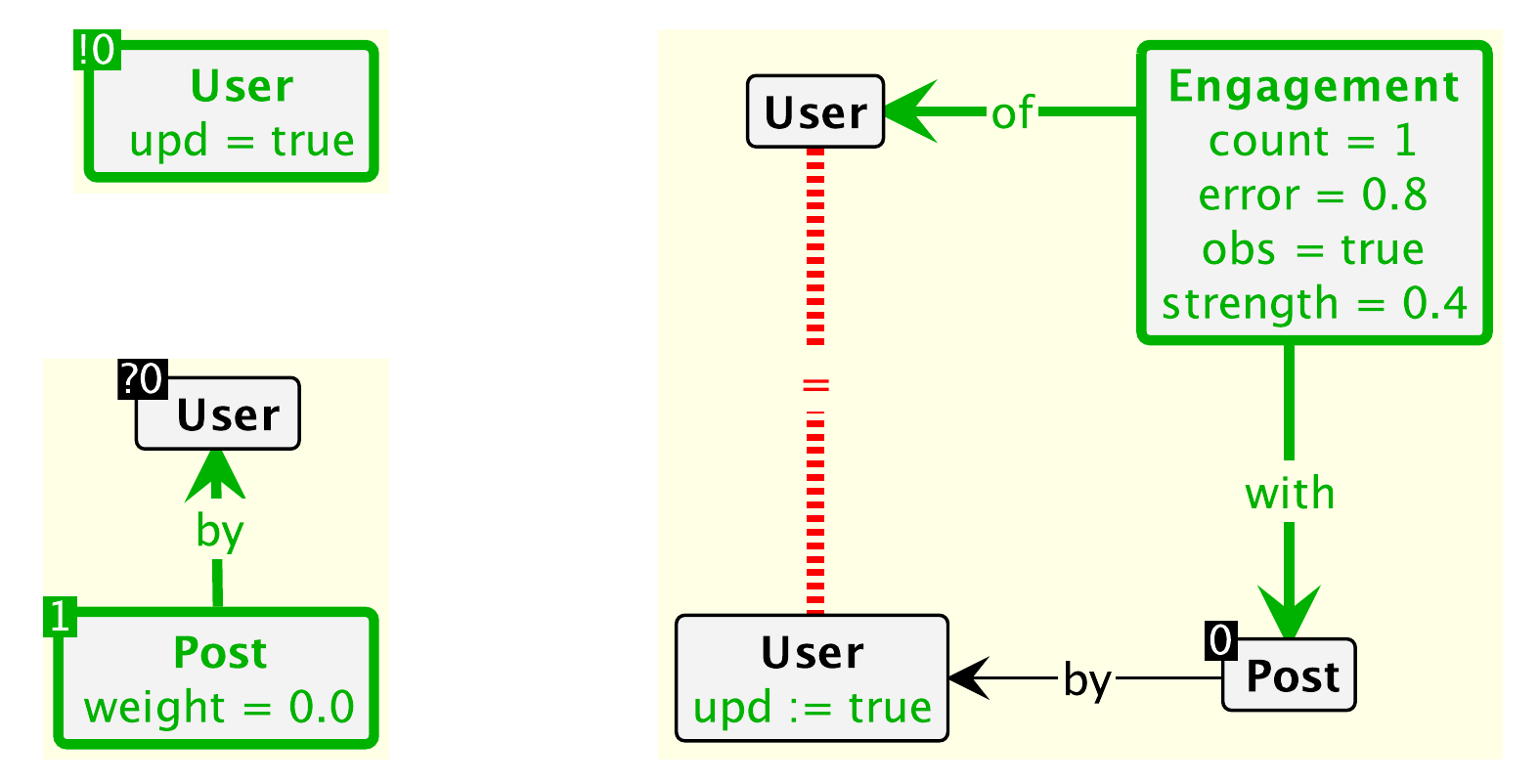}{0.4}{Rules for dynamic updates: \emph{newUser}, \emph{newPost}, and \emph{newEngagement04}.}

The rules come in two flavours: dynamic update rules to create users, posts and engagements, and rules for training and inference. Update rules are shown in Fig.~\ref{fig:dynamic} in Groove notation. As usual, elements in thin solid black outline represent read access, dashed blue elements are deleted, and thick green elements are created, while nodes and edges in dotted red represent negative conditions. The dotted red line labelled ``='' means that the linked nodes are not equal.
Rules to create engagements exist in three versions to create engagements of strengths $0.2, 0.4$ and $0.8$ (but only the rule for $0.4$ is shown). When creating a new engagement, the user who authored the post has their \emph{upd} attribute set to \emph{true} to trigger retraining.

\abb{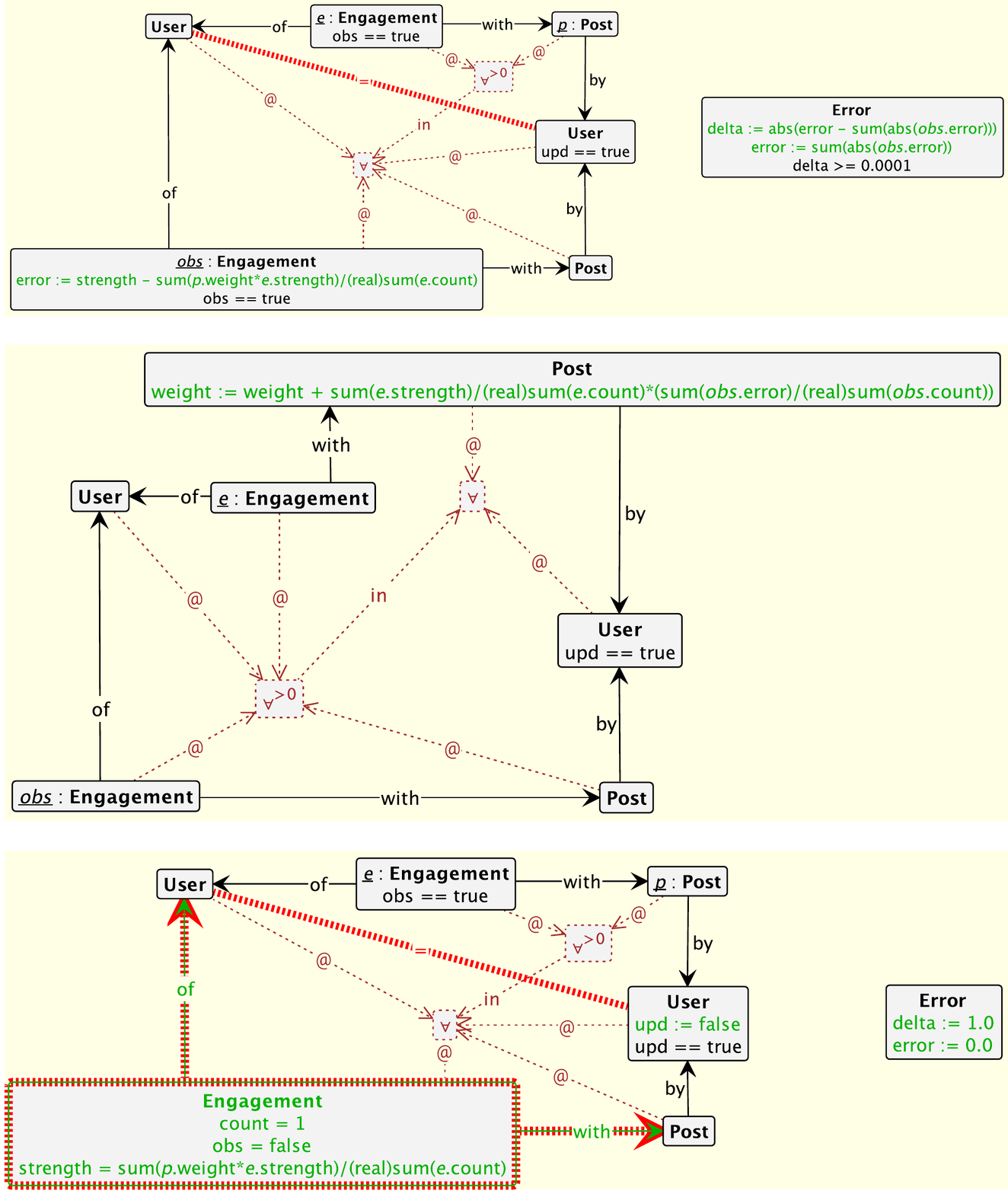}{0.62}{Top to bottom: Rules \emph{error, delta} and \emph{infer}.}


The training and inference rules are shown in Fig.~\ref{fig:train+infer}. Nodes labelled  $\forall$ or $\forall^>$ represent nested multi rules, $@$-labelled edges associating nodes to their nesting levels and $in$ edges showing quantifier nesting.
The \emph{infer} rule at the bottom calculates the expected strengths of new engagements as weighted sums of the strengths of all observed engagements of the same user with posts by a second user whose \emph{upd} attribute is \emph{true} (and then changes to \emph{false}). The red-green dotted outline says that an Engagement node is only created if there is not already one, combining node creation with a negative condition.
The assignment shown in green inside that Engagement node realises the following formula.
\begin{equation}
strength = \frac{\sum_{m \in M}(p(m).weight * e(m).strength)}{\mbox{card }M} \quad \label{eq:strength}
\end{equation}
$M$ is the set of matches of the multi pattern represented by the non-empty quantification node $\forall^>$, with $e(m)$ and $p(m)$ referencing the Engagement and Post nodes of match $m \in M$.
This is nested in another $\forall$ node to apply the inference to all eligible Engagement nodes at once. Recall that the weights of Post nodes show how well a post represents the average post by that user.

The \emph{error} rule in the top computes the inference error as the difference between the observed value of strength and the inferred value. 
This is typical of semi-supervised learning in GNNs where some of the nodes represent training samples while others are inferred. Structure and inference formula are similar to the \emph{infer} rule. We use an Error node with a global \emph{error} attribute to control the training cycle. This is calculated as the sum of absolute \emph{error} values of all observed engagements, and training stops when the \emph{delta}, the difference of global errors between this and the previous cycle is below $0.0001$.
The \emph{delta} rule in the middle of Fig.~\ref{fig:train+infer} implements gradient descent backward propagation, updating the weights of the posts by the product of the average error and strengths of the user’s engagements' with this post. 


We use a control program to iterate training and inference, and to simulate the insertion of newly observed data during the runtime of the model.
The program starts with a training cycle through \emph{error} and \emph{delta} for as long as possible, i.e., until the delta value falls below the threshold. Inference consists of one application of the \emph{infer} rule:
{\small
\begin{verbatim}
function training(){
   alap{error; delta;}
}
function inference(){
   infer;
}
function update(){
   node u; newUser(out u);
   node p; newPost(u, out p); 
   newEngagement02(p); newEngagement04(p); 
}
training; inference;
update;
training; inference;
\end{verbatim}
}

All rules use nested universal quantification over engagements, so a single application per rule covers the entire graph.
The sample dynamic update rules called by the \emph{update} function create a new user and post and two engagements. Rule parameters ensure that new posts, engagements and users are linked. Then we run the training and inference again, demonstrating the possibility of incremental updates and repeated training to reflect the interleaving of real-world changes and consequent data updates with the operation of the GReNN model. 

\begin{figure}
    \centering
    \includegraphics[width=\textwidth]{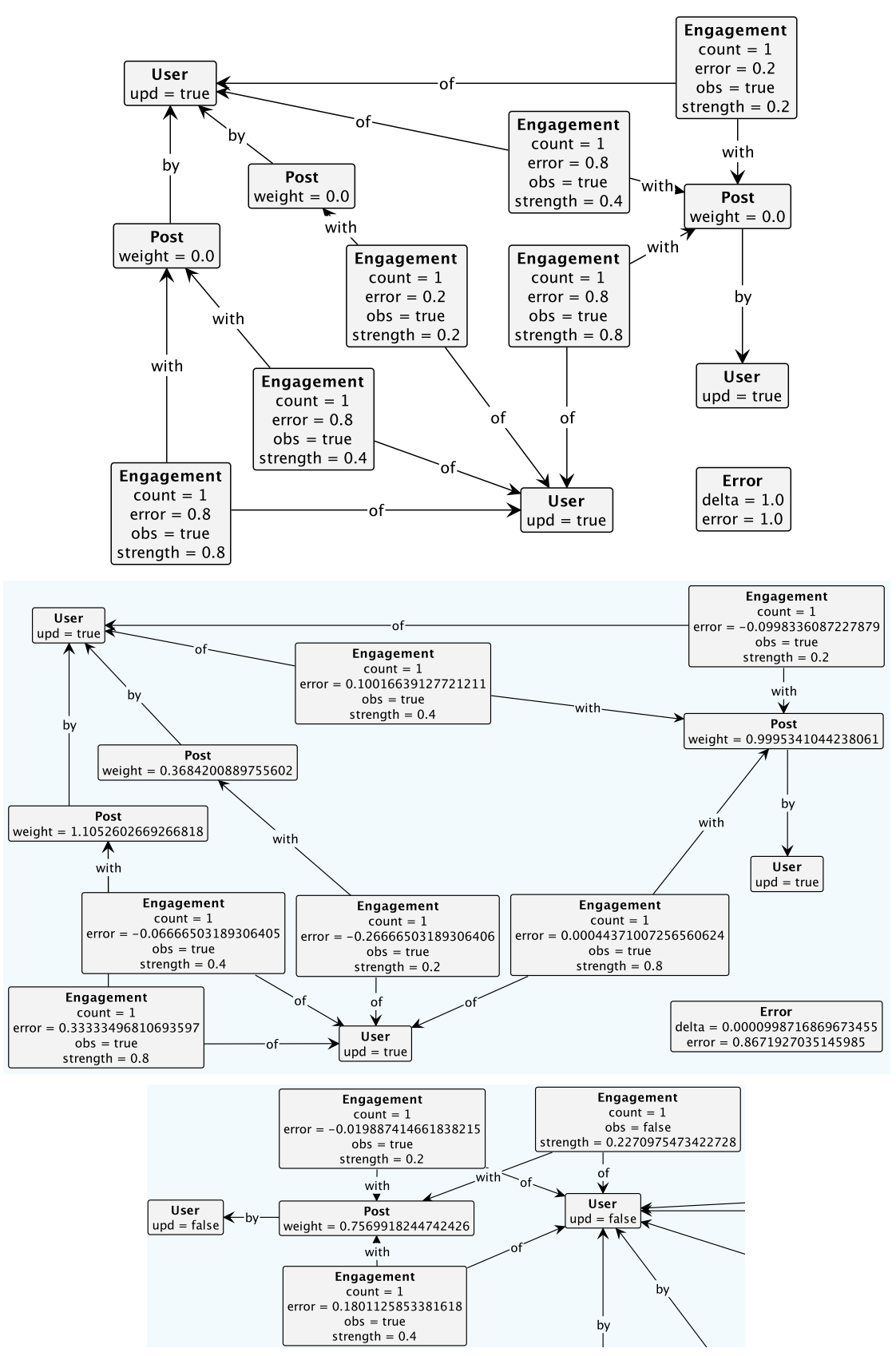}
    \caption{Top to bottom: start, intermediate and final graphs.}
    \label{fig:graphs}
\end{figure}

The start graph at the top of the Fig.~\ref{fig:graphs} has three users, three posts and six engagements, all representing ``observed'' data that can be used to train post weights. An intermediate graph, after the first training  cycle, is shown in the middle, with final global error and delta values, and updated post weights.  
The final graph extends this by, among other things, the structure at the bottom, where the user on the right of the fragment is the same as the one in the top left of the intermediate graph. In the final graph, it has two newly observed and one newly inferred engagement with a new post by the new user on the left.




\section{Discussion: GReNN vs GNN}

The purpose of this exercise was not to demonstrate how existing GNN approaches can be realised as graph rewrite systems, but to adopt principles of GNNs, such as the use of graphs to represent data for training and inference, to enable semi-supervised learning by labelling parts of the graph with real-world observations, and to employ the graph structure to direct training and inference. 

However, our model goes beyond mainstream GNNs in two important aspects. First, the graphs employed are heterogeneous (typed) directed graphs rather than homogeneous (untyped) undirected ones. This allows us to operate directly on an application-oriented data model such as a knowledge graph~\cite{kg-book} of a social media network, rather than an encoding. Knowledge graphs are used for data integration, graph-based analytics and machine learning~\cite{Schad2021}.

“Message-passing type GNNs, also called Message Passing Neural Networks (MPNN), propagate node features by exchanging information between adjacent nodes. A typical MPNN architecture has several propagation layers, where each node is updated based on the aggregation of its neighbours’ features”~\cite{michaelbeyond2022}. MPNNs realise a supervised machine learning algorithm on graph-structured data by utilising the message-passing algorithm~\cite{DBLP:journals/corr/GilmerSRVD17}. Notable variants of GNNs include Convolutional NNs~\cite{DBLP:conf/nips/DuvenaudMABHAA15}, Gated GNNs~\cite{li2017gated} and Deep Tensor NNs~\cite{Sch_tt_2017}. 

We relate our model to the concepts and terminology from~\cite{DBLP:journals/corr/GilmerSRVD17},
which describes the operation of MPNNs on undirected graphs $G$ by message passing, update and readout steps. Given node features $x_v$ and edge features $e_{vu}$, message passing updates hidden states $h^t_v$ at each node $v$ for $T$ steps according to message functions $M_t$ and vertex update functions $U_t$ using messages 
\[m^{t+1}_v = \sum\limits_{u\in N(v)} M_t(h^t_v,h^t_u,e_{vu})\] 
where $N(v)$ is the set of neighbours of $v$ and updates 
\[h^{t+1}_v = U_t(h^t_v,m^{t+1}_v).\]
After step $T$, readout $R$ computes an output vector for the graph.
Hence messages aggregated from nodes' neighbourhoods are used in updates to compute new states. $M_t$ and $U_t$ can be learned by neural networks. Often this means that $M_t$ computes a weighted sum of node and edge features, with weights trained by backpropagation, while $U_t$ implements activation thresholds or controls the rate of change by weighing residual states against massages. 

Our Eq.~(\ref{eq:strength}) can be seen as a combination of $M_t$ and $U_t$. The update function is the identity since no activation or balancing between residual and new features is required, and the message is a weighted average of engagement strengths with posts by the same user. However, the analogy is incomplete because our graphs are directed and not homogeneous, and the strength of engagement is a feature of an ``engaged with post by'' relation between users rather than a user node. Indeed ``one could also learn edge features in an MPNN by introducing hidden states for all edges in the graph \dots and updating them analogously'' \cite{DBLP:journals/corr/GilmerSRVD17}.
The relational nature of engagement is the reason why we only use a single step of forward propagation: All ``engagements with posts'' between the same two users are accessible locally in one step.
A complex pattern realises the relation, so we need a sum over the multi pattern to compute the average. Hence graph patterns allow querying heterogeneous graphs where their homogeneous encodings are processed by navigating along edges. Apart from supporting a direct representation of the data, graph patterns increase the expressive power of GNNs also on homogeneous graphs~\cite{bronstein_2022b}. 

Like weights and features, the graph structure is usually represented by matrices, such as  adjacency or incidence and degree matrices.
%
If the graph is undirected, its adjacency matrix is symmetrical. If the graph is weighted, it contains real-valued edge weights. 
A GReNN model supports directed edges, class heterogeneity and incremental updates to process dynamic application data in its original form without encoding. As of now, most GNN approaches only support undirected and homogenous graphs, but there is a trend towards more general models: In~\cite{HAN2022100201}, the authors present an example of a directed MPNN and use it in conjunction with a graph attention network. In~\cite{DBLP:journals/corr/GilmerSRVD17} the authors state that “It is trivial to extend the formalism to directed multigraphs”. 
%
Multi-class approaches are utilised in IoT applications such as~\cite{anjum_ikram_hill_antonopoulos_liu_sotiriadis_2013}. Emerging support for heterogeneous graph data includes~\cite{9802746,DBLP:journals/corr/abs-2109-00711} which present a new algorithm called Explicit Message-Passing Heterogeneous Graph Neural Network (EMP). While GReNNs support both directed and heterogeneous data at the model level, such approaches may provide a platform to implement our models efficiently.

Incremental forward and back propagation in GNNs can  substantially save time and resources~\cite{DBLP:journals/corr/abs-2006-14422} by reducing computations to nodes affected by updates. This is especially useful for big-data applications, where batch processing is not affordable and efficient. NNs and GNNs do not easily support data updates. 
Hence, the second generalisation in our approach is to  model the dynamic updates of the graph and control specifically where retraining  is required. This includes adding new nodes, an operation that would break most matrix-based representations by changing the dimensions of the matrices involved. 
 
GReNNs reimagine GNNs free of considerations of efficient representation of and computations on graphs. This is appropriate from a modelling perspective, but  unlikely to scale to large graphs.
To support an engineering approach to GNN development based on graph rewriting, a study of the possible implementations of such models in mainstream GNNs is required. First experiments show that it is possible in principle to map a model such as ours to an undirected homogeneous graph by interpreting the edges of that simpler graph as paths or patterns in our model. This is a reduction that GNN developers have to perform when they present their application data as input to a GNN, but we believe that by studying this systematically we can find standard mappings from a typed attributed graph model into graphs supported by suitable GNN technology.

The suggestion that graph rewriting can serve as a semantic foundation for GNNs requires mappings in the opposite direction, taking the various GNN approaches and modelling them as GReNNs. Then, such embeddings could be used to compare the different graph types, inference or training rules. The theory of graph grammars with its language hierarchies and corresponding rule formats may well play a role here.
GNNs can be based on attributed directed or undirected, simple or multi, homogeneous or typed, binary, hyper or hierarchical graphs, with little support for reuse of algorithms or implementations~\cite{sanchez-lengeling_reif_pearce_wiltschko_2021}. The categorical theory and potentially generic implementation of graph rewriting~\cite{DBLP:conf/gg/BrownPHF22} can be a model to unify such diverse approaches.       

\newcommand{\bibdir}{bibsRH}

\bibliographystyle{splncs03}
\bibliography{grenn,references,LITVS,greco,\bibdir/abbrev,\bibdir/articles,\bibdir/edition,\bibdir/informal,\bibdir/inprocs,\bibdir/tutorials,\bibdir/ag-engels,\bibdir/reiko,\bibdir/global,\bibdir/all}

\end{document}
\appendix

\section{Applications and Examples}

\subsection{Service-oriented Systems and other Distributed Architectures}

See discussion above. Services are interfaces to components. The relation between client and service is represented by an adapter. Dynamic binding of services is proof search in a suitable tile logic.  

One could also look at P2P or publish subscribe architectures. 

\subsection{Triple Graph Grammars and Graph-based Applications}

%
%

\end{document}